\documentclass[11pt]{article}

\usepackage[preprint]{acl}

\usepackage{times}
\usepackage{latexsym}

\usepackage[T1]{fontenc}

\usepackage[utf8]{inputenc}

\usepackage{microtype}

\usepackage{inconsolata}

\usepackage{graphicx}

\usepackage{booktabs}
\usepackage{multirow}
\usepackage[table]{xcolor}

\usepackage{xcolor}
\usepackage{tcolorbox}
\tcbuselibrary{skins, breakable}

\newcounter{example}

\newtcolorbox{examplebox}[1]{
  breakable,
  colback=white,
  colframe=black!60,
  boxrule=0.6pt,
  arc=2pt,
  left=8pt,
  right=8pt,
  top=6pt,
  bottom=6pt,
  fonttitle=\bfseries,
  coltitle=white,
  colbacktitle=black!65,
  before upper={\refstepcounter{example}},
  title={#1} 
}

\title{RelayGen: Intra-Generation Model Switching for Efficient Reasoning}

\author{
 \textbf{Jiwon Song\textsuperscript{1}},
 \textbf{Yoongon Kim\textsuperscript{1}},
 \textbf{Jae-Joon Kim\textsuperscript{1}}
\\
 \textsuperscript{1}Seoul National University
\\
\texttt{\{jiwon.song,yoon\_g\_kim,kimjaejoon\}@snu.ac.kr} \\
\url{https://github.com/jiwonsong-dev/RelayGen}
}

\begin{document}
\maketitle
\begin{abstract}
Large reasoning models (LRMs) achieve strong performance on complex reasoning tasks by generating long, multi-step reasoning trajectories, but inference-time scaling incurs substantial deployment cost.
A key challenge is that generation difficulty varies within a single output, whereas existing efficiency-oriented approaches either ignore this intra-generation variation or rely on supervised token-level routing with high system complexity. 
We present \textbf{RelayGen}, a training-free, segment-level runtime model switching framework that exploits difficulty variation in long-form reasoning. Through offline analysis of generation uncertainty using token probability margins, we show that coarse-grained segment-level control is sufficient to capture difficulty transitions within a reasoning trajectory.
RelayGen identifies model-specific switch cues that signal transitions to lower-difficulty segments and dynamically delegates their continuation to a smaller model, while preserving high-difficulty reasoning on the large model.
Across multiple reasoning benchmarks, RelayGen substantially reduces inference latency while preserving most of the accuracy of large models.
When combined with speculative decoding, RelayGen achieves up to 2.2$\times$ end-to-end speedup with less than 2\% accuracy degradation, without requiring additional training or learned routing components.
\end{abstract}

\section{Introduction}

The emergence of large reasoning models (LRMs)~\cite{o1,r1} has enabled language models to solve complex reasoning-intensive problems that were previously beyond reach. 
Recent progress in this area has been driven primarily by inference-time scaling, where substantially larger models are paired with longer generation trajectories, rather than by fundamental architectural changes.
As a result, models with 32B parameters or more have become the de facto standard for long-form reasoning tasks~\cite{qwq,qwen3,glm4}.

While this scaling trend improves reasoning accuracy, it significantly increases inference cost, making generation efficiency a central challenge for deploying LRMs in practice~\cite{stopoverthinking,efficientsurvey}.
In real-world settings, the computational overhead of long reasoning trajectories often dominates overall deployment cost, limiting the scalability of LRMs.

\begin{figure}[t]
  \centering
  \includegraphics[width=\linewidth]{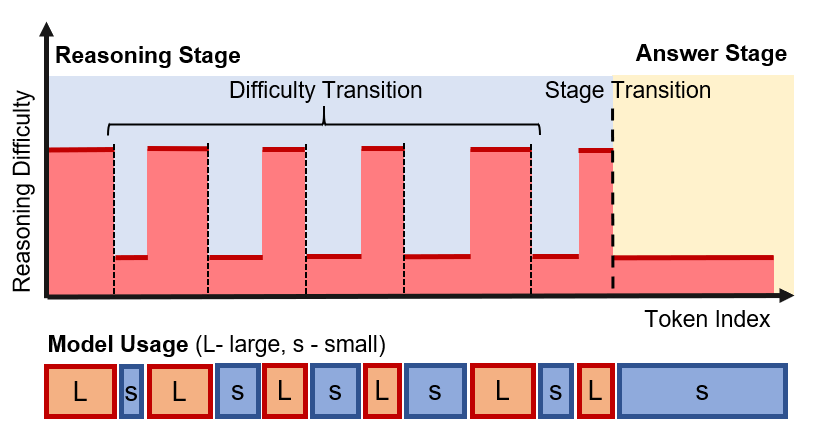}
  \vspace{-20pt}
  \caption{\textbf{RelayGen overview.} Long reasoning generation exhibits difficulty variation within a single output, enabling segment-level runtime model switching.}
  \vspace{-12pt}
  \label{fig:overview}
\end{figure}

This raises a natural question: \emph{do all parts of a long reasoning output require a large model?}
In practice, reasoning trajectories are heterogeneous, often interleaving high-difficulty reasoning with lower-difficulty continuation or consolidation.
As illustrated in Figure~\ref{fig:overview}, such difficulty variation arises within a single generation, suggesting opportunities for dynamic model allocation.

These observations suggest that long reasoning generation is inherently difficulty-heterogeneous rather than uniform.
However, existing efficiency-oriented approaches differ substantially in how they characterize and exploit such heterogeneity, often making trade-offs between practical deployability and the granularity at which difficulty variation is modeled within a single generation.

To improve inference efficiency, prior work has explored combining smaller, more efficient models with large models during inference.
Input-level routing methods~\cite{chen2023frugalgpt,ong2024routellm, zhang2025avengers} represent the simplest form of this approach, selecting one model per input and assigning it to the entire generation.
By construction, such methods treat all parts of an output uniformly and therefore cannot account for difficulty variation within a single long reasoning generation, which limits their applicability to LRMs.

More recent approaches attempt to account for intra-generation difficulty variation, primarily differing in the granularity at which routing decisions are made during decoding.
Token-level routing approaches~\cite{r2r,rstitch} query a separately trained router before generating each token, enabling difficulty-aware decisions at every decoding step.
While such token-level analysis provides a principled mechanism for identifying challenging positions during reasoning, it requires training and integrating an online routing model, introducing additional supervision requirements and system complexity that hinder practical deployment.

In contrast, part-level or step-level switching approaches~\cite{specthink} make routing decisions over larger units within an output. These methods often leverage lexical or structural cues to approximate difficulty, but tend to rely on heuristic criteria rather than a detailed analysis of model-specific reasoning behavior.
As a result, it remains unclear how well the selected segments align with actual difficulty across different models and tasks.

Consequently, existing approaches struggle to exploit difficulty variation within a single long reasoning output.
To address this gap, we analyze difficulty fluctuations along long reasoning generation trajectories of LRMs and observe that different parts of an output exhibit systematically different difficulty levels.
In particular, reflective, paraphrasing, and post-reasoning continuation segments often exhibit lower uncertainty than core reasoning segments.

Based on this observation, we propose \textbf{RelayGen}, a training-free, segment-level runtime model switching framework. 
RelayGen allocates large-model capacity to high-difficulty reasoning segments and hands off subsequent low-difficulty segments to a smaller model once a transition is detected. 
Rather than relying on supervised routing, RelayGen uses empirically grounded transition cues that are identified through difficulty analysis to guide segment-level model switching during generation.

Our key contribution is a simple, training-free routing mechanism grounded in empirical difficulty analysis, demonstrating that coarse-grained, segment-level control is sufficient to capture difficulty variation in long-form reasoning.
This finding challenges the prevailing reliance on fine-grained, learned routing for efficient and accurate inference with large reasoning models.
By operating at segment boundaries, RelayGen enables difficulty-aware inference that remains fully composable with speculative decoding, a property that is fundamentally incompatible with token-level routing schemes.

Our contributions are summarized as follows:
\begin{itemize}
  \setlength\itemsep{0em}
  \item We provide an empirical analysis of difficulty variation within long reasoning trajectories of large reasoning models.
  \item We propose RelayGen, a training-free, segment-level runtime model switching framework guided by empirically validated difficulty transitions.
  \item On AIME 2025 benchmark, RelayGen combines with speculative decoding to achieve up to 2.2$\times$ latency reduction with less than 2\% accuracy degradation.
\end{itemize}

\section{Related Works}

A line of work has explored improving inference efficiency by combining smaller models with more capable large models, delegating less demanding parts of generation to the former.

\paragraph{Input-level routing.}
Early approaches primarily operate at the input level~\cite{chen2023frugalgpt,ong2024routellm}, selecting a single model per request and assigning it to the entire generation.
Such input-level routing methods have been widely adopted for general LLM inference and have also been extended to large reasoning models~\cite{cprouter,routetoreason}.
However, by construction, they assign a uniform model throughout the generation and therefore cannot account for intra-generation difficulty variation that arises during the course of long reasoning generation.

\paragraph{Token-level routing.}
More fine-grained approaches attempt to delegate generation within a single output.
Token-level routing methods such as R2R~\cite{r2r} and R-Stitch~\cite{rstitch} selectively invoke a large model at token positions estimated to be difficult.
However, they rely on the assumption that difficulty can be localized at isolated token positions, often requiring trained routing models and additional supervision in practice.

\begin{figure}[t]
  \centering
  \includegraphics[width=0.85\linewidth]{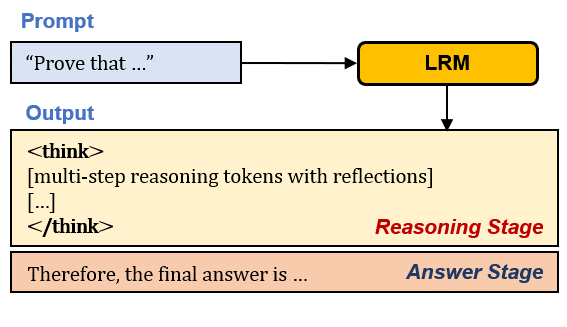}
  \vspace{-12pt}
  \caption{Given a prompt, an LRM generates an output that naturally decomposes into a long reasoning stage and a subsequent answer stage, explicitly separated by special boundary tokens (e.g., \texttt{<think>}, \texttt{</think>})}
  \vspace{-12pt}
  \label{fig:structure}
\end{figure}

\paragraph{Segment- or step-level switching.}
Another class of approaches operates at a coarser granularity by switching models over multi-token segments within an output.
Speculative Thinking~\cite{specthink} exemplifies this direction by dividing reasoning into steps using predefined delimiters and assigning steps based on heuristic cue tokens.
These cues are manually specified and applied in a model-agnostic manner, without being derived from an analysis of their actual impact on reasoning difficulty.
As a result, although such methods demonstrate that step-level intervention can be effective for long-form reasoning, the selected segments may not consistently align with regions of genuinely high or low difficulty for a given model and task distribution, which can lead to accuracy degradation in challenging reasoning settings.

\paragraph{Training-based switching.}
Related to these efforts, SplitReason~\cite{splitreason} proposes a training-based approach that learns a model-internal policy to decide when to hand off generation between models via special control tokens.
As a result, its design introduces a distinct accuracy–efficiency trade-off compared to training-free, runtime model allocation approaches based on externally observed generation dynamics.

\paragraph{Speculative decoding.}
In parallel, speculative decoding~\cite{specdec} represents a complementary line of work for reducing inference latency by jointly using small and large models.
In this paradigm, a small model~\cite{bild} or a learned speculator module~\cite{eagle3,medusa,deepseekmtp} proposes candidate token sequences that are subsequently verified by a large model, enabling lossless acceleration.
Speculative decoding is orthogonal to routing or switching methods and can therefore be combined with them to further accelerate generation.
However, because speculative decoding relies on predicting future tokens, it is difficult to be combined with token-level routing, while remaining naturally compatible with coarser-grained switching.

\section{Difficulty Variation in LRM Generation}
\label{sec:difficulty_variation}

\subsection{Structure of Long Reasoning Outputs}

As illustrated in Figure~\ref{fig:structure}, LRMs generate outputs that are explicitly structured into a reasoning stage followed by a subsequent answer stage, typically delimited by special tokens such as \texttt{<think>} and \texttt{</think>}.
During the reasoning stage, the model performs multi-step reasoning by exploring intermediate hypotheses and frequently revisiting earlier conclusions through reflection and self-checking.
The answer stage then produces the final response presented to the user, consolidating and summarizing the conclusions established during reasoning.

This explicit structure indicates that long reasoning outputs are not monolithic, but instead consist of parts with distinct generation characteristics.
In the following sections, we analyze how generation difficulty varies across these different parts.

\begin{figure}[t]
  \centering
  \includegraphics[width=1.0\linewidth]{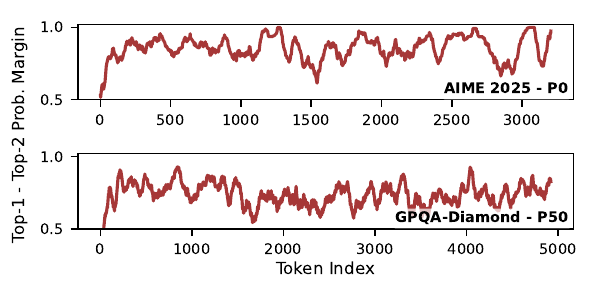}
  \vspace{-24pt}
  \caption{Probability margin trajectories of Qwen3-32B across long reasoning examples.}
  \vspace{-12pt}
  \label{fig:margin_trajectory}
\end{figure}

\subsection{Observed Difficulty Transition Signals \textit{during} Reasoning}
\label{sec:during_reasoning}

To quantify generation difficulty, we define the \emph{probability margin} as the difference between the probabilities of the top-1 and top-2 tokens.

Given the output logits $\mathbf{z}_t$ at generation step $t$, let $\mathbf{p}_t = \mathrm{softmax}(\mathbf{z}_t)$ denote the resulting token probability distribution.
The probability margin is defined as
\[
m_t = p_{t,(1)} - p_{t,(2)}
\]
where $p_{t,(1)}$ and $p_{t,(2)}$ denote the largest and second-largest probabilities, respectively.
Intuitively, a larger margin indicates that the model more decisively prefers its most likely next token over competing alternatives, whereas a smaller margin reflects higher uncertainty.

All margin statistics in this section are computed from Qwen3-32B reasoning traces on AIME 2025~\cite{aime25} and GPQA-Diamond~\cite{rein2024gpqa}.
\begin{figure}[t]
  \centering
  \includegraphics[width=1.0\linewidth]{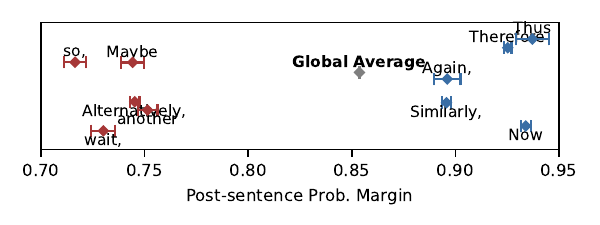}
  \vspace{-24pt}
  \caption{Post-sentence probability margin following representative discourse-level cues.}
  \vspace{-12pt}
  \label{fig:margin_impact}
\end{figure}
Figure~\ref{fig:margin_trajectory} shows that the probability margin fluctuates substantially over long reasoning trajectories, indicating non-uniform generation difficulty.

Here, we use the term \textit{discourse-level cues} to refer to lexical markers that signal shifts in the flow of reasoning discourse, including repair-oriented cues (e.g., \texttt{wait}, \texttt{but}) and conclusion-oriented cues (e.g., \texttt{therefore}, \texttt{so}).
Such cues are often associated with transitions from active reasoning to reflective or checking phases, during which the model elaborates on previously established conclusions.
To examine whether such transitions are reflected in generation uncertainty, we analyze how discourse-level cues correlate with probability margin dynamics.
Specifically, for each occurrence of a discourse-level cue, we compute \emph{post-sentence probability margin}, the average probability margin over the remainder of the sentence in which the cue appears.
The complete list of evaluated cues is provided in Appendix~\ref{appendix:discourse_level_cues}.

Figure~\ref{fig:margin_impact} shows that different discourse-level cues are followed by systematically different post-sentence probability margins, reflecting variation in subsequent generation difficulty.
For example, cues such as \texttt{Thus} are frequently followed by concluding or summarizing statements with higher post-sentence margins, whereas semantically similar cues such as \texttt{so} do not consistently exhibit the same behavior and may even be associated with lower margins.
This variability indicates that cue effects cannot be reliably predicted by simple heuristics and depend on model-specific generation dynamics, motivating empirical, model-specific analysis.

\subsection{Low-Difficulty Generation \textit{after} Reasoning}

\begin{table}[t]
\centering
\small
\caption{Answer-stage consistency when delegating answer generation from Qwen3-32B to Qwen3-0.6B.}
\vspace{-2pt}
\renewcommand{\arraystretch}{0.90}
\setlength{\tabcolsep}{8pt}
\begin{tabular}{c c c}
\toprule
Total Samples & Matches & Matching Rate \\
\midrule
728 & 727 & 99.86\% \\
\bottomrule
\end{tabular}
\label{tab:answer_consistency}
\vspace{-20pt}
\end{table}

The answer stage following the reasoning stage is typically easier to generate, as it does not involve independent reasoning but instead realizes conclusions established during the preceding stage in a condensed form.
Conditioned on the key-value cache of the reasoning stage, answer generation primarily involves summarization and formatting rather than exploration of new reasoning paths.
Notably, despite its lower reasoning difficulty, the answer stage is conditioned on the entire preceding reasoning context, resulting in long effective key-value cache.
As a result, answer stage tokens often incur the highest per-token attention cost during generation, even though their semantic difficulty is low.

This structure suggests that the answer stage may not require the full capacity of a large model, as its generation is heavily constrained by previously decoded reasoning tokens.
To examine this hypothesis, we conduct a controlled handoff experiment in which a small model generates the answer stage while being conditioned on reasoning traces produced by a large model.
Specifically, we generate full reasoning traces on problems from MATH500~\cite{math500}, AIME 2025, and GPQA-Diamond using Qwen3-32B, and then delegate the subsequent answer generation to Qwen3-0.6B.
Although Qwen3-0.6B is substantially smaller and weaker when used alone, we observe almost no answer flipping when only the answer stage is delegated to the smaller model.

Table~\ref{tab:answer_consistency} reports the consistency of the resulting answers compared to those produced entirely by the large model.
Out of 728 evaluated samples, only a single answer differs, corresponding to a matching rate of 99.86\%.
This result indicates that answer generation is highly stable under model handoff when conditioned on preceding reasoning, in sharp contrast to the substantial difficulty fluctuation observed within the reasoning stage itself.

\section{RelayGen}
\subsection{Overview of RelayGen}

\begin{figure}[t]
  \centering
  \includegraphics[width=1.0\linewidth]{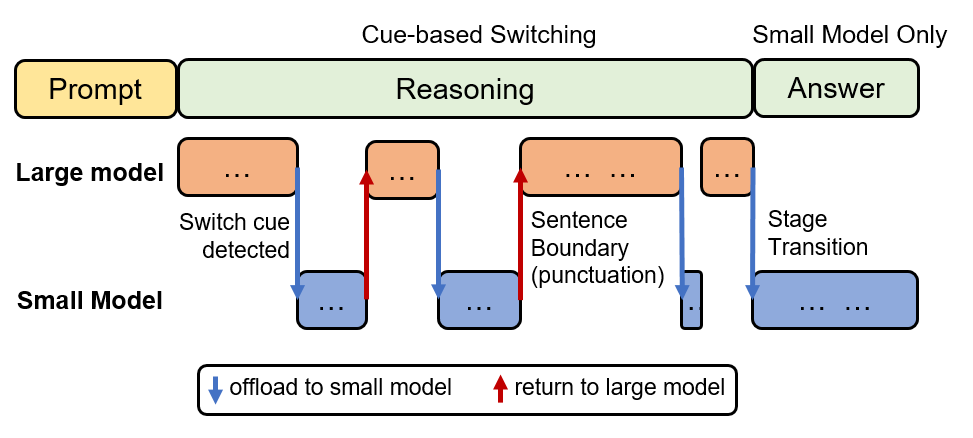}
  \vspace{-20pt}
  \caption{Runtime model switching semantics in RelayGen.}
  \vspace{-6mm}
  \label{fig:mechanism}
\end{figure}

RelayGen is a training-free, runtime framework for model switching during long reasoning generation.
Instead of assigning an entire output to a single model, RelayGen dynamically allocates model capacity within a single generation, matching model size to local generation difficulty.

RelayGen is motivated by empirical findings in Section~\ref{sec:difficulty_variation}, which show that
(i) generation difficulty varies substantially within a single reasoning trajectory,
(ii) discourse-level cues correlate with transitions between higher- and lower-difficulty regions, and
(iii) post-reasoning answer generation remains highly stable despite high per-token attention cost.
These observations suggest that different segments of a single generation benefit from different model capacities.

As illustrated in Figure~\ref{fig:mechanism}, RelayGen implements segment-level runtime model switching.
During the reasoning stage, generation is primarily performed by a large model to preserve accuracy.
When a selected switch cue is encountered, subsequent continuation segments are temporarily delegated to a smaller model.
After the transition from reasoning to the answer stage, the remainder of the output is generated entirely by the small model.

RelayGen operates entirely at runtime and requires neither additional training nor auxiliary routing components such as learned routers or control modules.

\subsection{Switch Cue Selection}
\label{sec:switch_cue_selection}

RelayGen does not directly use the full set of discourse-level cues analyzed in Section~\ref{sec:during_reasoning} for runtime switching.
Although many discourse-level cues are correlated with transitions in generation difficulty, not all such cues consistently lead to lower-uncertainty continuation.
In particular, some cues are associated with increased uncertainty and further exploration, making them unsuitable as indicators of low-difficulty segments.

To identify reliable switch cues, we perform an offline selection procedure based on \emph{post-sentence probability margin}, which is defined in Section~\ref{sec:during_reasoning}.
Importantly, this procedure does not involve optimization, learning, or parameter updates of any kind; it simply profiles generation statistics from fixed, pre-trained models.
Using an existing held-out dataset as a calibration set, we first generate reasoning traces using the large model.
For each occurrence of a candidate discourse-level cue in these traces, we compute the average probability margin over the remainder of the sentence, from the cue position to the next sentence boundary.

We then compare this post-sentence margin against the global average probability margin computed across all token positions in the calibration traces.
A discourse-level cue is selected as a switch cue if its post-sentence margin is higher than the global average by at least one standard error, indicating that the corresponding continuation is consistently easier for the model to generate.
Cues that do not satisfy this criterion, including those associated with decreased or inconsistent margins, are excluded from the switch cue set.

Importantly, this selection procedure does not involve learning a routing model or introducing additional supervision. 
Although it relies on an offline calibration set, the procedure is strictly training-free and only uses empirical generation statistics extracted from existing reasoning traces.
The resulting switch cue set is fixed at deployment time and used during runtime generation through simple token matching, as described in Section~\ref{sec:runtime_switching}. 
More details on switch cue selection are provided in Appendix~\ref{appendix:switch_cue_selection}.

\begin{table*}[t]
\centering
\small
\renewcommand{\arraystretch}{0.95}
\setlength{\tabcolsep}{8pt}
\caption{Pass@1 comparison of RelayGen and baselines. Results are reported for large–small model pairs, denoted as Large / Small.}
\vspace{-4pt}
\scalebox{0.95}{
\begin{tabular}{l ccc ccc}
\toprule
\multirow{2}{*}{Method} 
& \multicolumn{3}{c}{Qwen3-32 / 1.7B} 
& \multicolumn{3}{c}{R1-Distill-Qwen-32B / R1-Distill-Qwen-1.5B} \\
\cmidrule(lr){2-4} \cmidrule(lr){5-7}
& MATH500 & AIME 2025 & GPQA-Diamond
& MATH500 & AIME 2025 & GPQA-Diamond \\
\midrule
\rowcolor{gray!10}
Small Model
& 88.60 & 31.67 & 37.33
& 86.10 & 28.33 & 30.30 \\
\rowcolor{gray!10}
Large Model
& 95.27 & 70.00 & 64.58
& 93.50 & 53.30 & 60.61 \\
Spec.\ Think.
& 91.35 & 40.83 & 41.29
& 89.40 & 30.00 & 42.68 \\
R2R
& \underline{94.30} & \underline{62.50} & \underline{61.62}
& \underline{89.65} & \textbf{52.50} & \underline{48.99} \\
\rowcolor{blue!8} 
RelayGen
& \textbf{94.80} & \textbf{68.33} & \textbf{63.64}
& \textbf{91.70} & \underline{50.00} & \textbf{56.82}\\
\bottomrule
\end{tabular}
}
\label{tab:accuracy}
\end{table*}

\subsection{Runtime Switching Procedure}
\label{sec:runtime_switching}

RelayGen implements runtime model switching using standard generation control mechanisms, without relying on learned routers or online difficulty estimation.
As illustrated in Figure~\ref{fig:mechanism}, switching is realized by configuring model-specific stop conditions and issuing successive generation requests.
This design deliberately favors coarse-grained, externally observable control signals over fine-grained per-token routing, enabling simple and robust integration with existing inference systems.
RelayGen is built on top of vLLM and leverages its support for prefix caching and OpenAI-compatible generation APIs.

In RelayGen, generation begins with the large model.
Its generation request includes stop tokens corresponding to the selected switch cues, as well as the special token \texttt{</think>} that marks the transition from the reasoning stage to the answer stage.
When a switch cue is generated during reasoning, generation is temporarily halted and subsequent tokens are delegated to the small model.
After \texttt{</think>} is generated, the remainder of the output is generated entirely by the small model.

The small model uses sentence-ending punctuation as stop tokens.
Upon reaching a sentence boundary, control returns to the large model if reasoning is still ongoing, enabling sentence-bounded offloading that may occur multiple times within a single reasoning trajectory.
After the transition to the answer stage, the small model completes generation without further handoff.

Despite repeated bidirectional switching during reasoning, RelayGen does not require re-prefilling the entire prompt at each switch.
Leveraging vLLM~\cite{vllm}’s prefix caching, only newly generated tokens unseen by the target model are prefetched, keeping switching overhead minimal in practice.

Because RelayGen performs switching at coarse-grained segment boundaries rather than at every decoding step, it is naturally compatible with speculative decoding methods such as Eagle-3 and Multi-Token Prediction~\cite{eagle3,deepseekmtp}.
Speculative decoding can be applied whenever the large model is active without interfering with RelayGen’s switching logic, whereas token-level routing disrupts the draft--verify process, making effective composition with speculative decoding challenging.
We provide further analysis of this granularity mismatch in Appendix~\ref{appendix:token_level}.

Overall, RelayGen features runtime model switching that preserves composability without introducing significant routing overhead.

\section{Experiments}

\subsection{Experimental Setup}

\paragraph{Models.}
We evaluate RelayGen using two representative large--small model pairs.
For the Qwen3~\cite{qwen3} family, we use Qwen3-32B and Qwen3-1.7B, and for the R1-Distill~\cite{r1} family, we use R1-Distill-Qwen-32B and R1-Distill-Qwen-1.5B.
For all models, the maximum generation length is set to 32{,}768 tokens, with temperature 0.6 and top-$p$ sampling at 0.95.
For Qwen3 models, we additionally set top-$k$ to 20.
All sampling hyperparameters follow the recommended settings of each model family to ensure fair and stable comparison.

\begin{table*}[t]
\centering
\small
\caption{Inference speedup from single-model inference with large model and large-model utilization across different generation acceleration methods. Error bars denote standard deviation over problems
(5 problems, 5 runs per problem).}
\vspace{-4pt}
\renewcommand{\arraystretch}{1.0}
\setlength{\tabcolsep}{11pt}
\scalebox{0.9}{
\begin{tabular}{l c c c c c}
\toprule
Method & Eagle-3 & Spec.\ Think. & R2R & RelayGen & RelayGen + Eagle-3 \\
\midrule
Speedup ($\times$)
& $1.79${\scriptsize$\pm0.42$} & $2.21${\scriptsize$\pm0.95$} & $1.30${\scriptsize$\pm0.18$} & $1.29${\scriptsize$\pm0.20$} & $2.20${\scriptsize$\pm0.21$} \\
Large-Model Utilization (\%)
& $100.00${\scriptsize$\pm0.00$} & $25.54${\scriptsize$\pm6.82$} & $19.27${\scriptsize$\pm3.33$} & $69.80${\scriptsize$\pm3.62$} & $69.49${\scriptsize$\pm3.02$} \\
\bottomrule
\end{tabular}
}
\label{tab:speedup}
\end{table*}

\paragraph{Baselines.}
We compare RelayGen against single-model generation, where a single model produces the entire output, as well as mid-generation switching baselines.
These include R2R~\cite{r2r}, which employs a learned token-level router to select a model at each decoding step, and Speculative Thinking~\cite{specthink}, which performs step-level switching based on predefined lexical cues.

\paragraph{Datasets.}
We evaluate all methods on multiple reasoning benchmarks spanning different domains and difficulty levels.
For mathematical reasoning, we use AIME 2025~\cite{aime25} and MATH500~\cite{math500}, which exhibit substantially different difficulty levels.
To assess generality beyond mathematics, we additionally report results on GPQA-Diamond~\cite{rein2024gpqa}, which focuses on graduate-level scientific reasoning.

For all datasets, we generate four outputs per problem and report pass@1.

\paragraph{Implementation Details.}
RelayGen is implemented on top of vLLM~\cite{vllm} (version 0.13.0) using its OpenAI-compatible API.
Separate inference engines are instantiated for the large and small models.
Runtime model switching and prefix reuse are handled using standard generation controls and vLLM's built-in prefix caching mechanism, as described in Section~\ref{sec:runtime_switching}.

For offline calibration to derive model-pair-specific switch cues, we use the AMC 2023 dataset~\cite{amc2023} consisting of 40 problems.
For each problem, the large model generates four independent reasoning traces, resulting in a total of 160 calibration samples.

\subsection{Accuracy Evaluation}
\label{sec:accuracy}

Table~\ref{tab:accuracy} reports pass@1 on three reasoning benchmarks.
We compare RelayGen against single-model generation and mid-generation switching baselines, including Speculative Thinking and R2R, with the large and small models serving as upper and lower bounds, respectively.

Across both model pairs, RelayGen substantially improves accuracy over the small-only baseline while preserving most of the large model’s performance.
RelayGen consistently outperforms Speculative Thinking on all benchmarks.
Although Speculative Thinking also relies on lexical cues to guide mid-generation switching, it makes switching decisions without an explicit analysis of generation difficulty and operates at a coarse, step-level granularity.
As a result, large portions of reasoning are delegated to the small model even when difficulty remains high, leading to significant accuracy degradation.

Compared to R2R, which employs a learned token-level router, RelayGen achieves accuracy that is comparable overall, while exhibiting different strengths across benchmarks.
For the Qwen3 model pair, RelayGen outperforms R2R across all three benchmarks, including a notable improvement on AIME 2025.
For the R1-Distill model pair, RelayGen achieves higher accuracy on MATH500 and GPQA-Diamond, while R2R performs slightly better on AIME 2025.
These results suggest that segment-level switching is sufficient to preserve most of the large model’s reasoning capability, particularly when difficulty variation aligns with discourse-level structure.

Overall, RelayGen occupies a favorable point in the accuracy–efficiency spectrum, retaining a large fraction of the large model’s performance while avoiding the supervised training and fine-grained routing complexity required by token-level methods such as R2R.

\subsection{Inference Latency and Composability}
\label{sec:latency}

We evaluate inference latency using the Qwen3-32B / 1.7B model pair on the AIME 2025 dataset.
We randomly sample five problems and generate each problem five times, reporting the average end-to-end generation latency.
All experiments are conducted on two NVIDIA A100 80GB SXM GPUs under identical decoding configurations.
We report normalized speedup relative to the large model, along with the fraction of tokens generated by the large model, to characterize the latency--accuracy trade-off.

Table~\ref{tab:speedup} summarizes inference speedup and large-model utilization across different generation acceleration methods.
Speculative Thinking attains 2.21$\times$ of speedup compared to large-model-only baseline, primarily due to its low large-model utilization and coarse-grained switching.
However, as shown in Section~\ref{sec:accuracy}, this comes at the cost of substantial accuracy degradation.

R2R achieves relatively moderate speedup despite using large model for merely 19.3\% of tokens.
This indicates that its potential compute savings are largely offset by routing and model-switching overhead, as R2R invokes a neural router at every decoding step, disrupting continuous generation.

RelayGen exhibits inherent speedup comparable to R2R when applied alone, despite retaining substantially higher large-model utilization.
This efficiency stems from segment-level switching, which avoids the overhead of per-token routing decisions while preserving the large model for high-difficulty reasoning segments.
As a result, RelayGen achieves a more favorable accuracy–latency trade-off, as evidenced in Section~\ref{sec:accuracy}.

Crucially, RelayGen is fully compatible with speculative decoding.
When combined with Eagle-3, RelayGen achieves a final speedup up to 2.20$\times$, substantially outperforming R2R while remaining entirely training-free.
Moreover, the combined approach yields an additional 1.22$\times$ speedup over Eagle-3 alone, demonstrating that RelayGen complements speculative decoding by selectively reducing end-to-end latency without aggressive early switching.

Overall, these results indicate that RelayGen prioritizes stable latency reduction through selective offloading, retaining most generation on the large model while remaining composable with existing inference acceleration techniques.

\begin{table}[t]
\centering
\small
\caption{Effect of switch cue selection on pass@1.}
\vspace{-4pt}
\renewcommand{\arraystretch}{0.85}
\setlength{\tabcolsep}{8pt}
\begin{tabular}{lcc}
\toprule
Cue usage & AIME 2025 & GPQA-Diamond \\
\midrule
\rowcolor{blue!5}
Selected & 68.33 & 63.64 \\
All candidates & 60.00 & 57.32 \\
\bottomrule
\end{tabular}
\label{tab:cue_selection_ablation}
\vspace{-4pt}
\end{table}

\begin{figure}[t]
  \centering
  \includegraphics[width=0.98\linewidth]{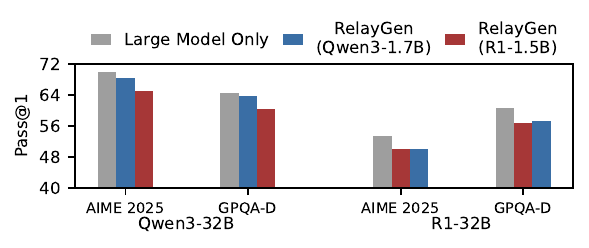}
  \vspace{-10pt}
  \caption{Pass@1 under different small-model choices for each large model, shown relative to the large-model-only baseline.}
  \vspace{-12pt}
  \label{fig:model_change_ablation}
\end{figure}

\subsection{Ablation Studies}
\label{sec:exp_ablation}

\paragraph{Effect of Switch Cue Selection.} 
To isolate the impact of switch cue selection, we conduct an ablation study on the AIME 2025 and GPQA-Diamond with Qwen3-32B and 1.7B model pair.
We compare RelayGen with selected switch cues against a variant that activates all discourse-level cue candidates as switching triggers. 
As shown in Table~\ref{tab:cue_selection_ablation}, using all candidate cues consistently reduces pass@1 for both benchmarks. 
These results indicate that indiscriminate use of discourse-level cues leads to suboptimal switching decisions, highlighting the importance of accurate cue selection for effective model switching.


\paragraph{Effect of Changing Model Pairs.}
We evaluate the robustness of RelayGen to model pair selection by testing heterogeneous model pairs on AIME 2025 and GPQA-Diamond.
Specifically, we consider Qwen3-32B / R1-Distill-Qwen-1.5B and R1-Distill-Qwen-32B / Qwen3-1.7B, with results shown in Figure~\ref{fig:model_change_ablation}.
We find that the performance drop for the Qwen3-32B / R1-Distill-Qwen-1.5B pair, relative to its large-model baseline, is comparable to that observed for the R1-Distill-Qwen-32B / R1-Distill-Qwen-1.5B pair.
This suggests that the degradation primarily stems from the limited capacity of the small model rather than incompatibility across model families.
Overall, RelayGen remains effective with heterogeneous model pairs when the small model has sufficient capacity.


\begin{table}[t]
\centering
\small
\caption{Effect of calibration set size on pass@1.}
\vspace{-4pt}
\renewcommand{\arraystretch}{0.85}
\setlength{\tabcolsep}{6pt}
\begin{tabular}{ccc}
\toprule
\# Calibration samples & AIME 2025 & GPQA-Diamond \\
\midrule
10  & 70.00 & 61.87 \\
40  & 71.67 & 61.87 \\
\rowcolor{blue!5}
160 & 68.33 & 63.64 \\
\bottomrule
\end{tabular}
\label{tab:calibration_size_ablation}
\end{table}

\paragraph{Sensitivity to Calibration Set Size.}
We study the sensitivity of RelayGen to the calibration set size using the Qwen3-32B / Qwen3-1.7B model pair, where our main experiments adopt 160 calibration samples.
As shown in Table~\ref{tab:calibration_size_ablation}, reducing the calibration set size to 40 or even 10 samples does not lead to significant performance degradation on either AIME 2025 or GPQA-Diamond.
In some cases, smaller calibration sets yield comparable or slightly higher pass@1 scores, suggesting that RelayGen is not overly sensitive to the exact choice of calibration size.
Overall, these results indicate that RelayGen remains effective even with very small calibration sets, reinforcing its practicality in training-free and low-overhead deployment scenarios.

\section{Conclusion}
In this work, we showed that generation difficulty within long reasoning trajectories of large reasoning models is inherently heterogeneous, and that not all parts of a single output require the full capacity of a large model.
Based on an offline empirical analysis of token-level uncertainty, we proposed RelayGen, a training-free, segment-level runtime model switching framework that dynamically allocates model capacity within a single generation without relying on learned routers or additional supervision.
Across multiple reasoning benchmarks, RelayGen preserves most of the accuracy of large models while substantially reducing inference latency. 
Moreover, RelayGen composes effectively with speculative decoding, achieving up to 2.2$\times$ end-to-end speedup.
More broadly, our results suggest that difficulty-aware inference for long-form reasoning can be effectively addressed through empirically grounded, coarse-grained runtime mechanisms, rather than fine-grained, supervised routing.

\section*{Limitations}

RelayGen targets settings where long-form reasoning is explicitly externalized, as in modern LRMs, and where generation difficulty varies over extended trajectories.
For tasks that do not require sustained reasoning or whose outputs lack clear long-sequence structure, the opportunity for effective segment-level model switching may be limited.

RelayGen also assumes a moderate capability gap between the large and small models.
If the small model lacks sufficient reasoning capacity even for relatively low-difficulty segments, offloading can degrade overall performance.

While our experiments focus on English reasoning data, the underlying principle—switching models at points of reduced reasoning difficulty—is not inherently language-specific.
We leave a systematic exploration of such multilingual extensions to future work.

\section{Ethical Considerations}

RelayGen is a training-free inference-time framework and does not introduce new model capabilities, data sources, or learning signals beyond those of the underlying models.
As such, it does not raise additional ethical concerns compared to standard deployment of large language models.
Any biases or risks are inherited from the base models, and RelayGen only controls runtime allocation between them.



\bibliography{custom}

\newpage

\onecolumn

\appendix

\section{Discourse-level Cues}
\label{appendix:discourse_level_cues}

This appendix specifies the set of discourse-level cues considered throughout the analysis in RelayGen.
We emphasize that this appendix only defines which tokens are treated as discourse-level cues and does not describe how switch cues are selected or used for model handoff, which is detailed separately in Appendix~\ref{appendix:switch_cue_selection}.

We define discourse-level cues as lexical tokens that do not introduce new problem-specific content, but instead signal structural transitions in a reasoning process, such as progression, reconsideration, or consolidation of intermediate conclusions.

The cue pool used in this work is constructed by starting from an existing list of reflection keywords introduced in~\cite{nowait}, which focuses on in-reasoning reflections, and extending it with additional tokens that explicitly signal result consolidation or summary transitions (e.g., \texttt{so}, \texttt{therefore}).
The motivation for this augmentation is to cover not only intermediate reasoning reflections, but also discourse signals that commonly precede conclusion formation in long-form reasoning outputs.

Importantly, the resulting cue pool is fixed prior to any model-pair-specific calibration and serves as a closed set of tokens considered in all subsequent analyses.

For clarity, we categorize the discourse-level cues according to their intended discourse function in a reasoning process.  
This categorization is descriptive and does not imply that all cues are suitable or effective as switch signals.

\begin{itemize}
    \item \textbf{Progression and Continuation Cues.}
    These tokens indicate forward movement within an ongoing reasoning chain or the continuation of the current line of thought.
    Typical examples include \texttt{now}, \texttt{then}, \texttt{next}, and \texttt{again}.

    \item \textbf{Reconsideration and Branching Cues.}
    These tokens signal a pause, correction, or exploration of an alternative reasoning path, often reflecting local uncertainty or revision.
    Examples include \texttt{wait}, \texttt{however}, \texttt{alternatively}, and \texttt{instead}.

    \item \textbf{Inference and Transition Cues.}
    These tokens mark a transition from intermediate reasoning steps to an inferred statement or implication.
    Representative examples include \texttt{thus}, \texttt{hence}, and \texttt{therefore}.

    \item \textbf{Result Consolidation and Summary Cues.}
    These tokens frequently precede the consolidation or summarization of conclusions, particularly near the end of a reasoning process.
    Examples include \texttt{so}, \texttt{therefore}, and related connective markers.
\end{itemize}

All discourse-level cues considered in this work fall into one of the above categories.
The complete cue list used in our analysis is summarized in Table~\ref{tab:discourse_cue_pool}.

\begin{table}[h]
\centering
\small
\renewcommand{\arraystretch}{1.0}
\setlength{\tabcolsep}{6pt}
\begin{tabular}{l l}
\toprule
\textbf{Category} & \textbf{Discourse-Level Cues (Canonical Form)} \\
\midrule
Progression / Continuation
& \texttt{now}, \texttt{then}, \texttt{next}, \texttt{again} \\
\addlinespace
Reconsideration / Branching
& \texttt{wait}, \texttt{however}, \texttt{alternatively}, \texttt{but}, \texttt{maybe}, \texttt{hmm}, \texttt{oh} \\
\addlinespace
Inference / Transition
& \texttt{thus}, \texttt{hence}, \texttt{therefore}, \texttt{similarly}, \texttt{specifically} \\
\addlinespace
Result Consolidation / Summary
& \texttt{so}, \texttt{therefore}, \texttt{check}, \texttt{double-check}, \texttt{verify} \\
\addlinespace
Reference / Enumeration
& \texttt{another}, \texttt{other}, \texttt{any} \\
\addlinespace
Discourse Acknowledgement
& \texttt{ah} \\
\bottomrule
\end{tabular}
\caption{Canonical discourse-level cue pool used for analysis.
For clarity, only canonical forms are shown; capitalization and punctuation variants are handled separately during tokenization.}
\label{tab:discourse_cue_pool}
\end{table}

\newpage

\section{Switch Cue Selection Details}
\label{appendix:switch_cue_selection}

This appendix describes how switch cues are selected from the discourse-level cue pool defined in Appendix~\ref{appendix:discourse_level_cues}.
We emphasize that being a discourse-level cue does not necessarily imply that a token consistently eases reasoning or leads to more stable continuation behavior.
The effect of a cue can vary substantially depending on model characteristics and learned representations, and even the direction of its impact may differ across models.
As a result, switch cues cannot be reliably chosen based on intuition alone and must instead be identified empirically.

\paragraph{Profiling Method.}
To this end, we apply the same analysis framework introduced in Section~\ref{sec:during_reasoning} to a set of reasoning traces generated on a calibration set.
For each discourse-level cue, we collect statistics over all occurrences of the cue in the generated traces.
Specifically, after a cue token appears, we measure the average probability margin from that token until the end of the sentence.
This post-sentence probability margin is then compared against the global average probability margin computed over all token positions.

We select, for each model pair, cues whose post-occurrence probability margin is, on average, higher than the global token-level average.

\paragraph{Model Pair Dependency.}
Switch cue selection is performed independently for each large--small model pair.
This reflects the fact that different models may respond differently to the same discourse token due to variations in architecture, scale, and training data, leading to differences in both cue sensitivity and the direction of its effect.

\paragraph{Calibration Setup.}
The profiling procedure is training-free and requires only a modest number of reasoning traces.
In our main experiments, we use the AMC 2023 dataset~\cite{amc2023} consisting of 40 problems.
For each problem, the large model generates four independent reasoning traces, resulting in a total of 160 calibration samples, which are reused for profiling under the small model.

To identify segments that are relatively easier for the small model to continue, we feed these traces into the small model and compute token-level probability margins, which are then aggregated at the discourse level for switch cue
selection.
As shown in the ablation study in Section~\ref{sec:exp_ablation}, this procedure remains robust even when substantially fewer calibration samples are used.


\paragraph{Calibration Overhead.}
RelayGen employs a lightweight offline calibration procedure that is executed once per large--small model pair.
Table~\ref{tab:calibration_overhead} reports the wall-clock time of this procedure measured on two NVIDIA A100 GPUs using the AMC 2023 dataset.
The calibration cost is dominated by reasoning trace generation with the large model, which takes approximately 80 minutes, followed by probability margin extraction and discourse-level aggregation under the small model, which require an additional 20 minutes.

Overall, the total one-time calibration cost is about 100 minutes and is incurred entirely offline, introducing no overhead during inference.
Moreover, since RelayGen is robust to smaller calibration sets, as demonstrated in our ablation study, this cost represents a conservative upper bound and can be significantly reduced in practice without degrading performance, as the calibration overhead scales with the number of calibration samples.

\begin{table}[h]
\centering
\small
\caption{Offline calibration overhead for RelayGen measured on two NVIDIA A100 GPUs
using the AMC 2023 dataset.}
\label{tab:calibration_overhead}
\renewcommand{\arraystretch}{1.05}
\setlength{\tabcolsep}{10pt}
\begin{tabular}{l c}
\toprule
\textbf{Calibration Stage} & \textbf{Wall-clock Time} \\
\midrule
Large-model reasoning trace generation & 80 minutes \\
Probability margin extraction and cue-level aggregation & 20 minutes \\
\midrule
Total calibration time (one-time, offline) & 100 minutes \\
\bottomrule
\end{tabular}
\end{table}

\paragraph{Final Switch Cue Sets.}
The resulting switch cue sets selected for each model pair are summarized in Table~\ref{tab:switch_cue_sets}.

\begin{table}[h]
\centering
\setlength{\tabcolsep}{8pt}
\renewcommand{\arraystretch}{1.0}
\resizebox{1.0\linewidth}{!}{
\begin{tabular}{l p{0.7\linewidth}}
\toprule
Large--Small Model Pair & Selected Switch Cues \\
\midrule
Qwen3-32B / Qwen3-1.7B\allowbreak~\cite{qwen3} &
\{ \texttt{"Oh,"}, \texttt{"another,"}, \texttt{"Thus"}, \texttt{"Now"}, \texttt{"Alternatively"}, \texttt{"alternatively,"},
\texttt{"Thus,"}, \texttt{"Therefore"}, \texttt{"similarly"}, \texttt{"similarly,"}, \texttt{"now"}, \texttt{"Again"},
\texttt{"specifically,"}, \texttt{"Again,"}, \texttt{"Similarly,"}, \texttt{"Now,"}, \texttt{"Specifically,"}, \texttt{"Hence"},
\texttt{"Similarly"}, \texttt{"Other"}, \texttt{"now,"}, \texttt{"hence"}, \texttt{"Specifically"}, \texttt{"So "},
\texttt{"Therefore,"}, \texttt{"Wait,"}, \texttt{"Also"}, \texttt{"So,"} \} \\

R1-Distill-Qwen-32B / R1-Distill-Qwen-1.5B~\cite{r1} &
\{ \texttt{"Wait"}, \texttt{"Thus"}, \texttt{"thus"}, \texttt{"similarly"}, \texttt{"Again,"}, \texttt{"Now"},
\texttt{"Therefore"}, \texttt{"hence"}, \texttt{"Hence,"}, \texttt{"Now,"}, \texttt{"Thus,"}, \texttt{"Oh,"},
\texttt{"Similarly,"}, \texttt{"Any"}, \texttt{"Therefore,"}, \texttt{"Alternatively,"}, \texttt{"now,"}, \texttt{"So,"},
\texttt{"now"}, \texttt{"verify"}, \texttt{"Specifically,"}, \texttt{"Alternatively"}, \texttt{"Ah,"}, \texttt{"wait"},
\texttt{"So "} \} \\
\bottomrule
\end{tabular}
}
\caption{Model-pair-specific switch cue sets selected offline from the discourse-level cue pool defined in Appendix~\ref{appendix:discourse_level_cues}.}
\label{tab:switch_cue_sets}
\end{table}

\newpage

\section{Granularity Mismatch between Token-Level Routing and Speculative Decoding}
\label{appendix:token_level}

Speculative decoding relies on a clear separation between a draft phase, in which a lightweight model proposes a contiguous block of tokens, and a verification phase, in which a larger model verifies these tokens in bulk.
This design amortizes the cost of large-model computation over many accepted tokens and is most effective when long draft spans can be formed.
In contrast, token-level routing strategies~\cite{r2r,rstitch} determine model usage at every generation step, requiring a routing decision before each token is generated.
As a result, the draft process is frequently interrupted, preventing the formation of long contiguous draft blocks and forcing verification to be repeatedly reset.
This mismatch in decision and verification granularity fundamentally limits the effectiveness of speculative decoding when combined with token-level routing.

Example~\ref{example:r2r} illustrates a generation trace produced by a token-level routing method, R2R~\cite{r2r}.
In the example, tokens generated by the large model are highlighted in \textcolor{red}{red}, while the remaining tokens are produced by the small model.
As shown, the large model is invoked only for sparse positions, often for a single token at a time, before control immediately returns to the small model.
This behavior reflects the core objective of token-level routing approaches: they aim to minimize large-model utilization to obtain speedups, often requiring additional machinery such as training an explicit router that decides model usage at every generation step.

\begin{examplebox}{Example 1: R2R Generation Example}
\label{example:r2r}
Okay, so I have this problem about Aya walking and stopping at a coffee shop.
Let me try to understand it step by step.

First, Aya walks 9 kilometers every morning.
When she walks at a constant speed of $s$ kilometers per hour,
the total time taken, including \textcolor{red}{$t$} minutes in the coffee shop, is 4 hours.
Then, when she increases her speed to $s + 2$ \textcolor{red}{km}/h,
the total time becomes 2 hours and 24 minutes,
\textcolor{red}{which is} 2\textcolor{red}{.}4 hours.

The question is asking, if she walks at $s + \frac{1}{2}$ km/h,
how many minutes will the walk take, including the $t$ minutes in the coffee shop.

Hmm, so \textcolor{red}{I need} to find the time it takes
when her speed is $s + \frac{\textcolor{red}{1}}{2}$ km/h.
\textcolor{red}{But} first, I think I need to find the value of $s$ and $t$.

Let me denote \textcolor{red}{the walking} time \textcolor{red}{without} the coffee shop
as $t_{\textcolor{red}{1}}$ and $t_2$
when she walks at $s$ and $s + 2$ km/h respectively.
\textcolor{red}{Then}, the total time is
$t_1 + t$ (\textcolor{red}{in} hours) \textcolor{red}{and} $t_{\textcolor{red}{2}} + t$ (in hours).

But wait, the problem says the total time including \textcolor{red}{$t$} minutes.
So, $t$ is \textcolor{red}{the} time spent in the coffee shop,
which is the same in both cases.
So, when she walks at $s$ km/h, the total time is 4 hours,
which \textcolor{red}{includes} $t$ minutes.
Similarly, when she walks at $s + 2$ km/h, the total time is
2 hours and 24 minutes, which is 2.4 hours,
and again includes $t$ minutes.

So, let me convert all the times into hours to make it easier.
The coffee shop time $t$ is in minutes, so I need to convert it to hours.
Since 1 hour is 60 minutes, $t$ minutes is $t/60$ hours.

So, \textcolor{red}{for} the first scenario:
Walking \textcolor{red}{time +} Coffee shop time = 4 hours  
Which is $(9 / s) + (t / 60) = 4$
\newline
For the second scenario:  
Walking time + Coffee shop time = 2.4 hours  
Which is $(9 / (s + 2)) + (t / 60) = 2.4$

So, now I have two equations:
\newline1) $9 / s + t / 60 = 4$  
\newline2) $9 / (s + 2) + t / 60 = 2.4$

I can \textcolor{red}{subtract} the second equation from the first
to eliminate $t / 60$.
\newline So, $(9 / s \textcolor{red}{-} 9 / (s + 2)) = 4 - 2.4 = 1.6$

\textcolor{red}{So}, $9 / s - 9 / (s + 2) = 1.6$

Let me compute \textcolor{red}{this}:
$\textcolor{red}{9} * (1 / s - 1 / (s + 2)) = 1.6$

Factor out $9$: $9*(1/2 - 1/(s+2))=1.6$

Compute $1 / s - 1 / (s + 2):
= (s + 2 - s) / [s (s + 2)]
= 2 / [s (s + 2)]$

\textcolor{red}{So}, $9 * (2 / [s (s + 2)]) = 1.6$ 

Simplify: $18 / [s (s + 2)] = \textcolor{red}{1.6}$

\textcolor{red}{Multiply} both sides by $s (s + 2)$: [...]

\end{examplebox}

However, this fine-grained routing leads to extremely short and fragmented large-model segments, which prevents effective composition with speculative decoding.
Speculative decoding relies on forming long contiguous draft spans so that the target model can verify many tokens in bulk and amortize the verification cost.
Under token-level routing traces such as Example~\ref{example:r2r}, verification spans collapse to one or two tokens per invocation, eliminating most of the amortization benefit while leaving switching and verification overheads dominant.

Moreover, even though the small model is used for the majority of tokens in token-level routing, the resulting speedup is often far from proportional, as we demonstrated in Section~\ref{sec:latency}.
This suggests that the overhead of making routing decisions at every generation step is non-negligible in practice.
In contrast, RelayGen avoids per-token routing decisions by switching at a coarser, segment-level granularity, enabling longer contiguous spans that compose naturally with speculative decoding.
As a result, RelayGen achieves comparable or greater speedups without aggressively suppressing large-model utilization or introducing per-step routing overhead.

\end{document}